\documentclass[10pt,twocolumn,letterpaper]{article}

\usepackage{iccv}
\usepackage{times}
\usepackage{epsfig}
\usepackage{graphicx}
\usepackage{amsmath}
\usepackage{amssymb}

\usepackage[ruled,vlined]{algorithm2e}
\usepackage{booktabs}
\usepackage{enumitem}
\usepackage{gensymb}
\usepackage{subcaption}

\DeclareMathOperator*{\argmin}{arg\,min}

\usepackage[pagebackref=true,breaklinks=true,letterpaper=true,colorlinks,bookmarks=false]{hyperref}

\iccvfinalcopy % *** Uncomment this line for the final submission

\def\iccvPaperID{} % *** Enter the ICCV Paper ID here

\ificcvfinal\fi

\begin{document}

%%%%%%%%% TITLE
\title{Camera View Adjustment Prediction for Improving Image Composition}

\author{
Yu-Chuan Su \qquad Raviteja Vemulapalli \qquad Ben Weiss \qquad Chun-Te Chu\\
Philip Andrew Mansfield \qquad Lior Shapira \qquad Colvin Pitts\\
Google Research
}

\maketitle
\ificcvfinal\thispagestyle{empty}\fi

%%%%%%%%% ABSTRACT
\begin{abstract}
Image composition plays an important role in the quality of a photo.
However, not every camera user possesses the knowledge and expertise required for capturing well-composed photos.
While post-capture cropping can improve the composition sometimes, it does not work in many common scenarios in which the photographer needs to adjust the camera view to capture the best shot.
To address this issue, we propose a deep learning-based approach that provides suggestions to the photographer on how to adjust the camera view before capturing.
By optimizing the composition before a photo is captured, our system helps photographers to capture better photos.
As there is no publicly-available data for this task,
we create a view adjustment dataset by repurposing existing image cropping datasets.
Furthermore, we propose a two-stage semi-supervised approach that utilizes both labeled and unlabeled images for training a view adjustment model.
Experiment results show that the proposed semi-supervised approach outperforms the corresponding supervised alternatives,
and our user study results show that the suggested view adjustment improves image composition $79\%$ of the time.
\end{abstract}

%%%%%%%%% BODY TEXT
\section{Introduction}

Image composition has a significant effect on the perception of an image.
While a good composition can help make a great picture out of the dullest subjects and plainest of environments, a bad composition can easily ruin a photograph despite how interesting the subject may be.
Unfortunately, a typical camera user may lack the knowledge and expertise required to capture images with great composition.

\begin{figure}
    \vspace{-5pt}
    \centering
    \includegraphics[width=\linewidth]{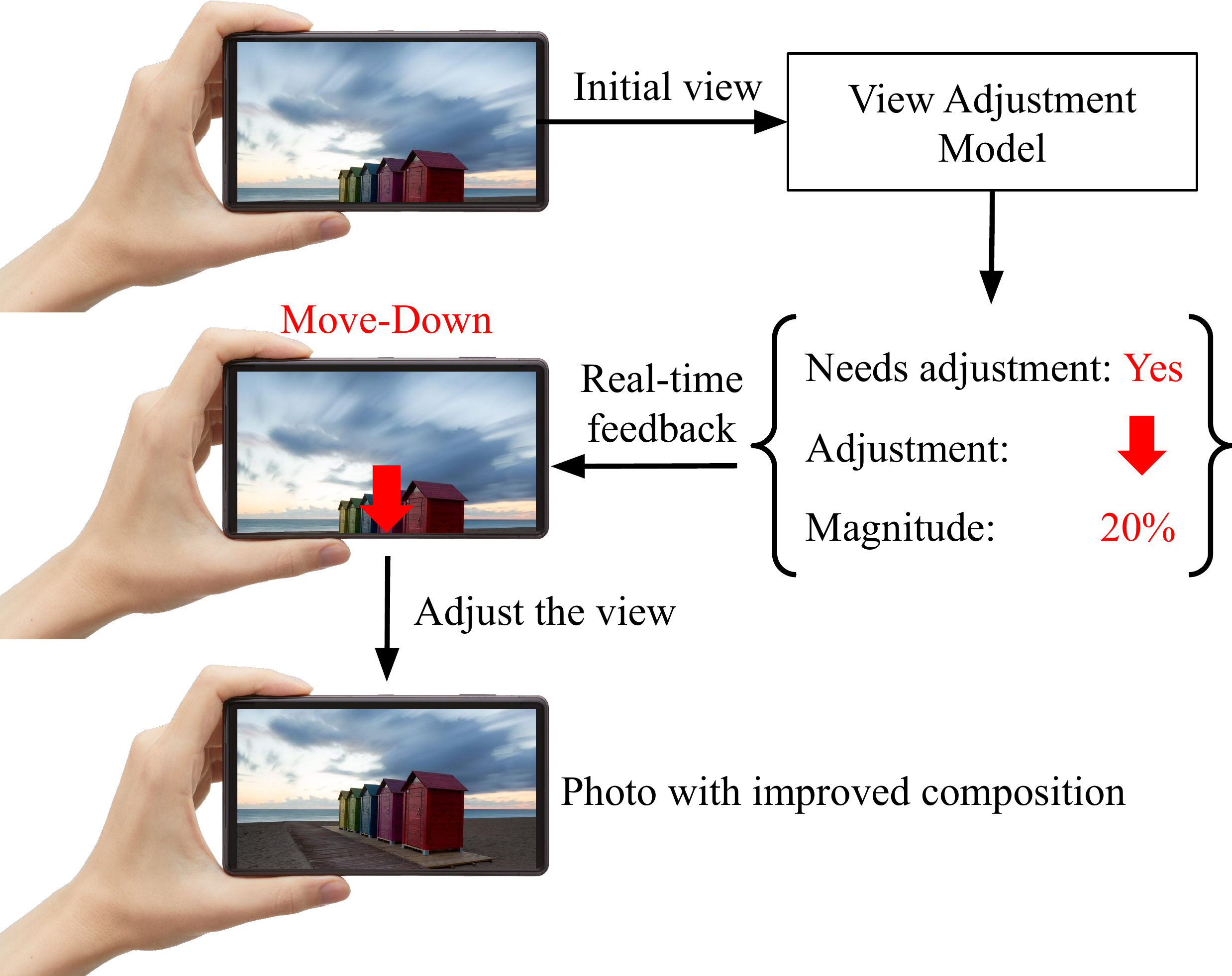}
    \vspace{-18pt}
    \caption{
        Our goal is to improve the composition of the captured photo by providing view adjustment suggestions when the user is composing the shot.
    }
    \vspace{-18pt}
    \label{fig:concept}
\end{figure}

A commonly used technique for improving image composition is image cropping,
and several existing works study how to crop images automatically~\cite{suh2003automatic,chen2003visual,stentiford2007attention,marchesotti2009framework,chen2016automatic,zhang2005auto,liu2010optimizing,fang2014automatic,nishiyama2009sensation,wang2017deep,deng2018aesthetic,chen-acmmm-2017,fang2017creatism,yan2013learning,chen-wacv2017,wei2018good,zhang2019deep,lu2019listwise,abs-1911-10492,zeng2020cropping}.
However, cropping works only in limited scenarios in which the best composition can be achieved by removal of certain portions of the image.
It is not suitable in many common scenarios where the photographer needs to adjust the camera view to get the best shot.
When evaluated on our view adjustment dataset, the crops predicted by the state-of-the-art GAIC cropping model~\cite{zeng2020cropping} have an average IoU of 0.61 with the groundtruth views, clearly indicating that cropping is not enough. In comparison, the proposed view adjustment model achieves a much higher IoU of 0.75.

While there are some existing rules for composing photos,  each rule is valid only for specific scenes, and requires detection of various low-level (leading lines, triangles, etc.) and high-level semantic (face/person, foreground objects) cues. Furthermore, it is non-trivial to determine which rule or combination of rules is applicable for a given scene.

\begin{figure}[t]
    \centering
    \includegraphics[width=\linewidth]{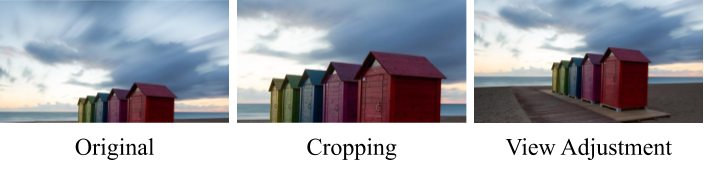}
    \vspace{-18pt}
    \caption{
        View adjustment enables more diverse modification to image composition, and can improve composition in scenarios where image cropping fails.
    }
    \vspace{-12pt}
    \label{fig:crop_vs_adjust}
\end{figure}

To address this problem,
we introduce a system that provides camera view adjustment suggestions to the photographer when they are composing the shot.
Given a view composed by the user,
our goal is to suggest a candidate view adjustment and its magnitude such that the photo captured after applying the adjustment will have a better composition (see Fig.~\ref{fig:concept}).
Specifically, we consider the following adjustments in this work:
\textit{horizontal (left or right)}, \textit{vertical (up or down)}, \textit{zoom (in or out)},
and \textit{rotation (clockwise or counter clockwise)} along the principal axis.
The adjustment magnitude is represented using a percentage of the image size for all adjustments except rotation, for which we use radians.
By adjusting the view prior to capture, we enable generic modifications to image composition.
Hence, our system can improve image composition in scenarios where cropping fails (see Fig.~\ref{fig:crop_vs_adjust}).
Note that this work focuses on static scenes,
or more specifically scenes where motion is not the subject (e.g.~portraits, nature and urban environment photography).
Suggesting view adjustments may not be suit dynamic (wildlife, sports, actions) scenes where lightning-quick decisions are required from the photographer.

To the best of our knowledge,
there is no publicly-available data for evaluating the performance of view adjustment prediction.
While a common practice is for human raters to annotate images with ground truth labels,
this is difficult for view adjustment because the results of adjustment are generally not available.
The raters need to infer how the adjustment affects the composition, which may be difficult without professional photography knowledge.
Instead, we create a new view adjustment dataset from existing image cropping datasets.
The idea is to convert view adjustments into operations on 2D image bounding boxes and use the best crop annotation as the target view for adjustment.
The proposed approach allows us to generate samples and view adjustment labels from image cropping datasets automatically without additional human labor.

We acknowledge that our view adjustment dataset is not ideal in several respects.
It ignores perspective distortions while adjusting the camera view,
and the adjustment magnitude could be limited depending on the ratios between the best crop and the uncropped image sizes. Despite these limitations,
our view adjustment dataset still provides a good starting point to evaluate composition-aware view adjustment prediction models.
Furthermore, we address these limitations by also evaluating the view adjustment model on $360\degree$ images,
which do not suffer from distortions or limited field-of-view (FOV).
Please refer to Sec.~\ref{sub:subjective} for details.

\begin{figure*}
    \vspace{-8pt}
    \centering
    \includegraphics[width=\linewidth]{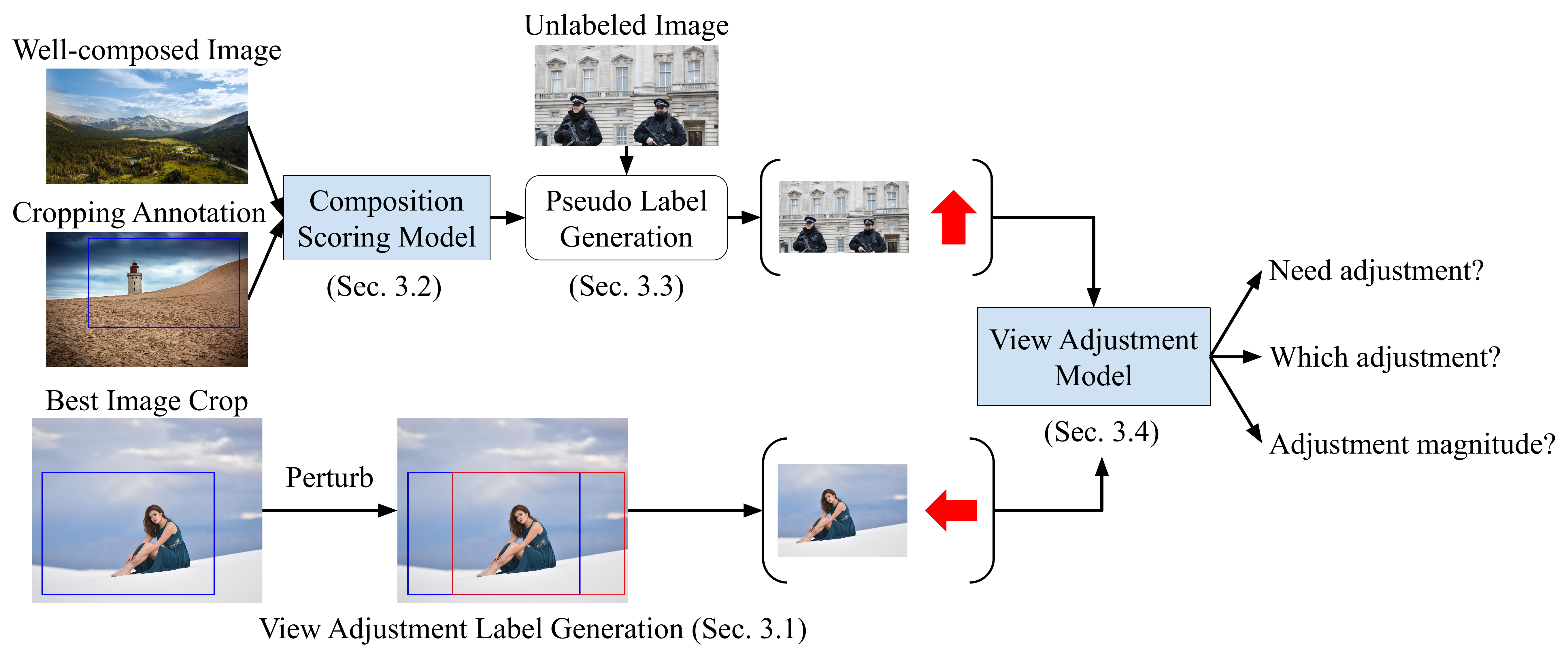}
    \vspace{-21pt}
    \caption{
        Our two-stage semi-supervsied approach leverages both labeled and unlabeled data to learn the view adjustment model.
        In the first stage, we learn a composition scoring model from both labeled data and unlabeled well-composed images.
        The scoring model is used to generate pseudo view adjustment labels for unlabeled images.
        These pseudo-labeled images along with real labeled images are used to train the view adjustment model in the second stage.
    }
    \vspace{-15pt}
    \label{fig:overall_approach}
\end{figure*}

Another limitation of our view adjustment dataset is that its amount and diversity is inherently limited by the cropping datasets.
Because state-of-the-art machine learning models typically require a large and diverse dataset to train well,
our view adjustment dataset may not be sufficient for training a good view adjustment model.
In light of this problem, we propose a two-stage training approach that makes use of additional unlabeled images.
See Fig.~\ref{fig:overall_approach}.
Our empirical results show that the additional unlabeled data is important for improving model performance.

We evaluate our approach on a view adjustment dataset consisting of 3,026 samples generated from 521 images from the FCDB~\cite{chen-wacv2017} and GAICD~\cite{zhang2019deep} datasets.
Quantitative results show that the proposed semi-supervised approach clearly outperforms the corresponding supervised alternatives,
and user study shows that the adjustments suggested by our model improve the composition $79\%$ of the time.

Our major contributions are as follows.
First, we formulate the problem of view adjustment prediction for improving image composition.
Second, we introduce a labeled dataset for evaluating view adjustment prediction models.
Finally, we propose a two-stage semi-supervised approach that leverages both labeled and unlabeled data to train the view adjustment model.
We show that the proposed semi-supervised approach outperforms the corresponding supervised alternative quantitatively and demonstrate the effectiveness of our model through user study.

\section{Related Work}
\paragraph{Image cropping}
Cropping is a widely used technique for changing image composition during post-processing,
and automatic image cropping algorithms aim to find the best crop within an image.
One common approach followed by existing methods is to select the candidate crops using a scoring function,
and the research focus has been on designing a good scoring function for cropping.
Existing works exploit saliency~\cite{suh2003automatic,chen2003visual,stentiford2007attention,marchesotti2009framework,chen2016automatic}, photography rules~\cite{zhang2005auto,liu2010optimizing,fang2014automatic},
or a data driven approach to learn the scoring function~\cite{nishiyama2009sensation,wang2017deep,deng2018aesthetic,chen-acmmm-2017,fang2017creatism,yan2013learning,chen-wacv2017,wei2018good,zhang2019deep,lu2019listwise,abs-1911-10492,zeng2020cropping,Li_2020_CVPR_Mutual}.
While early methods learn the cropping model using unsupervised data~\cite{chen-acmmm-2017,fang2017creatism} or data annotated for generic image aesthetic quality~\cite{nishiyama2009sensation,wang2017deep,deng2018aesthetic},
recent works show that a large scale annotated dataset designed specifically for cropping is essential for learning the state-of-the-art image cropping model~\cite{wei2018good,zhang2019deep,lu2019listwise,abs-1911-10492,zeng2020cropping}.
Instead of following the scoring function paradigm,
some works try to predict the target crop directly without generating and scoring candidate crops~\cite{li2018a2,li2019fast,lu2019end,Li_2020_CVPR,lu2020weakly}.

Our goal is not to produce the best image crop; it is to improve image composition while the photographer is composing the image.
Because view adjustment is a more general operation than image cropping, it may improve composition in scenarios where cropping is not suitable.
Furthermore, a view adjustment model needs to make a suggestion based on partial information,
while the target crop is fully visible to image cropping models.
This introduces unique challenges in view adjustment prediction for both data collection and modeling.

\paragraph{Photography recommendation} Prior works on photography recommendation provide various types of suggestions.
Some of them study person location recommendation~\cite{zhang2012icip,xu2014should,ma2014pose,ma2014icip,wang2015tist,rawat2018tmm}.
They take a scenic image as input and suggest where a person should stand within the image.
Others study photography at popular landmarks and suggest either a view enclosure in the input image or a geo-location for capturing a better photo~\cite{huang2012mm,yin2012crowdsourced,zhang2013location,yin2013socialized,rawat2015context,rawat2016clicksmart}.
These works utilize abundant web photos captured near the target landmark in order to provide landmark-specific suggestions.
Our problem differs from the above in that we provide view adjustment suggestions to the photographer.
Furthermore, our method is applicable to any photo instead of being restricted to specific image content or geo-locations.
Similar to our approach, Virtual Portraitist~\cite{hu2019virtualportraitist} utilizes weakly-labeled ``positives'' and suggests how to adjust the head pose to approximate positive samples, which can be considered as an inverse camera view adjustment.
However, it is designed specifically for learning head poses in selfies and is not applicable for other types of photos.
While some works also provide view recommendation in arbitrary photos~\cite{cheng2010learning,su2012tmm},
they assume the availability of a wide-angle image and find the target view within it.
Therefore, they require specialized hardware and is not applicable to any camera.
Finally,
some commercial systems provide photography recommendation in the form of suggested viewfinder center~\cite{camera51,samsung-shot-suggestion} but provide only a black box system.
To the best of our knowledge,
this is the first work that formulates and introduces an evaluation benchmark for the view adjustment prediction problem.

\vspace*{-0.05in}
\section{Approach}
\vspace*{-0.05in}
\label{sec:approach}
In this section,
we introduce our approach for learning view adjustment prediction.
We first formalize our problem and provide an overview of the two-stage training approach.
Next, we describe the two models and our pseudo-label generation process in more detail.

\subsection{Problem Formulation}
\label{sub:problem_formulation}
\vspace*{-0.05in}

Given an input image,
the goal of view adjustment is to answer the following questions:
1) whether there exists a candidate view adjustment that will improve the composition of the image,
2) if the composition can be improved, which candidate view adjustment can lead to the best composition,
and 3) given the candidate view adjustment, what is the appropriate magnitude for the adjustment.

To create a dataset with the view adjustment labels for the above questions,
we propose generating the sample images and ground truth labels from image cropping data.
Given an image and the best crop within the image,
we perturb the image crop by one of the candidate view adjustments with a random magnitude.
We take the perturbed crop as the image sample and use the inverse perturbation as the ground truth label for view adjustment.
We also use the best crop as the sample, where the label is that no adjustment is required.
The candidate adjustments considered in this work include \textit{horizontal (left or right)}, \textit{vertical (up or down)}, \textit{zoom (in or out)}, and \textit{rotation (clockwise or counter clockwise)} about the principal axis.
We focus on these eight adjustments because they can be presented in the camera viewfinder and executed by the photographer easily.
Note that rotation about yaw and pitch axes may be hard to distinguish from translation on a 2D display, and their effects may be similar depending on the scene.
Also, more complex adjustments can be achieved by performing multiple basic adjustments in a sequence.
We control the perturbation magnitude such that 1) the perturbed crop has sufficient overlap with the annotated crop,
and 2) the perturbed crop is in the original image.
See supp.~for data generation details.

While the proposed approach allows us to generate training samples for view adjustment from image cropping datasets,
the diversity of the generated data is controlled by the source cropping datasets.
To overcome this limitation, we propose a two-stage training approach that makes use of additional unlabeled images.
In the first stage, we learn a composition scoring model that rates the composition quality of a given image.
We then use this scoring model to generate pseudo view adjustment labels for unlabeled images.
This is achieved by simulating candidate view adjustments for a given image and selecting the adjustment that leads to the best scoring image as the pseudo label.
In the second stage, we train a view adjustment model using both the labeled images generated from cropping datasets and the additional pseudo-labeled images.
See Fig.~\ref{fig:overall_approach}.
The two-stage training approach exploits unlabeled data to increase the amount of training data and improve model performance in both stages.
Details are provided in the following sections.
Also see supp.~for the training pipeline summary.

\subsection{Composition Scoring Model}
\label{sec:composition_quality_model}

We implement the composition scoring model using a deep convolution neural network (CNN), denoted $M_c$.
The model takes an RGB image as input and produces a score in the range $[0, 1]$ as output.
Since the purpose of this model is to compare the results of different view adjustments, we train it using a pairwise ranking loss.
Let $(I_p, I_n)$ be two images of the same scene where $I_p$ has better composition than $I_n$.
The pairwise ranking loss encourages the score of $I_p$ to be more than the score of $I_n$ by a margin $\delta$:
\begin{equation}
\label{eq:ranking_loss}
    \max(0,\ \delta + M_{c}(I_n) - M_{c}(I_p)).
\end{equation}
To train this model, we use pairwise comparison data generated from both labeled image cropping data and unlabeled well-composed images.

\vspace*{-0.08in}
\paragraph{Labeled data}
We utilize image cropping datasets in the following formats to generate image pairs $(I_{p}, I_{n})$:
\begin{itemize}[leftmargin=*,label=$\bullet$,topsep=4pt]
    \setlength{\itemsep}{1pt}
    \setlength{\parskip}{1pt}
    \item \emph{Scored candidate crops}---in this format, each image comes with a pre-defined set of candidate crops along with their composition scores.
    In each training iteration, we choose $N$ crops from a random image and generate $\frac{N(N-1)}{2}$ pairs.
    $I_{p}$ and $I_{n}$ are determined for each pair by comparing the provided scores.
    We use the pairwise ranking loss averaged over these $\frac{N(N-1)}{2}$ pairs, denoted as $L_{sc}$, for training.
    \item \emph{Best crop annotation}---this format provides a best crop for each image.
    We take the best crop as $I_{p}$ and generate $I_{n}$ by randomly perturbing the bounding box of the best crop.
    We restrict the perturbation magnitude such that $I_n$ remains within the original image.
    See supp.~for the perturbation details.
    In each training step, we generate $K$ pairs from $K$ images and use the pairwise ranking loss averaged over these $K$ pairs, denoted as $L_{bc}$, for training.
\end{itemize}

\vspace*{-0.05in}
\paragraph{Unlabeled data}
Modern CNNs require a large and diverse dataset for training.
While recent image cropping datasets provide a large number of candidate crops~\cite{wei2018good,zhang2019deep,zeng2020cropping}, they usually come from a relatively small number of images.
To increase the amount and diversity of the training data,
we exploit unlabeled images following prior works in learning image aesthetic~\cite{chen-acmmm-2017,fang2017creatism}.
We first collect well-composed images from a photography website where the photos are contributed by experienced photographers\footnote{\href{https://unsplash.com/data}{unsplash.com}}.
Given a well-composed image,
we randomly perturb the image (by cropping, shifting, zooming-out, or rotation) to generate a new image with inferior composition.
We use the original image as $I_{p}$ and the perturbed image as $I_{n}$ to form an image pair.
See supp.~for perturbation details.
Unlike the perturbation in the labeled data, we allow the generated image to go beyond the original image.
Regions in $I_n$ that are not visible in $I_p$ are filled with zero pixels.

Our method for using unlabeled data differs from prior works as we use perturbations beyond cropping.
This is important because our goal is to provide more general view adjustment,
and empirical results show that using cropping alone leads to worse performance.
In each training step, we generate $P$ pairs from $P$ unlabeled images and use the ranking loss averaged over these $P$ pairs, denoted as $L_{wc}$, for training.
The total loss function used to train the composition scoring model is given by 
$L = L_{sc} + L_{bc} + L_{wc}$.

\vspace*{-0.05in}
\paragraph{Data augmentations}
Many of the pairs generated from unlabeled images will have zero-pixel borders in $I_n$ but not in $I_p$.
When trained on such pairs, the composition scoring model may just learn to score images based on the presence or absence of zero-pixel borders instead of focusing on the content.
To avoid this, we introduce three types of zero-pixel border augmentations:
\begin{itemize}[leftmargin=*,label=$\bullet$,topsep=4pt]
    \setlength{\itemsep}{1pt}
    \setlength{\parskip}{1pt}
    \item \emph{Shift borders} selects a certain percentage of columns on either the left or the right side of the image, and a certain percentage of rows at either the top or the bottom of the image.
    The pixel values in selected rows and columns are replaced with zeros.
    This simulates the zero-pixel borders introduced by shift perturbations.
    \item \emph{Zoom-out borders} selects a certain percentage of rows and columns on all sides and replaces the pixel values in these rows and columns with zeros.
    This simulates the zero-pixel borders introduced by zoom-out perturbation.
    \item \emph{Rotation borders} selects a random $\theta$ and applies $R_{-\theta} \circ R_{\theta}$ to the image, where $R_{\theta}$ is the rotation operator.
    This simulates the borders introduced by a rotation perturbation.
\end{itemize}
We randomly apply one of these augmentations to image $I_p$ in the pairs generated from unlabeled images so that the model learns to focus on the image content rather than zero-pixel borders.
We also add these augmentations to both $I_p$ and $I_n$ in the pairs generated from labeled data.

Our composition scoring model is motivated by existing cropping models~\cite{chen-acmmm-2017,wei2018good,zhang2019deep}.
The main difference is that existing cropping models are designed to compare image crops that are fully visible to the model.
In contrast, the goal of our model is to compare the results of different view adjustments where many of them are only partially visible.
Also, we combine both labeled and unlabeled data to train the model, while existing cropping models are trained with only one of them.

\begin{figure}
    \vspace{-3pt}
    \centering
    \includegraphics[width=\linewidth]{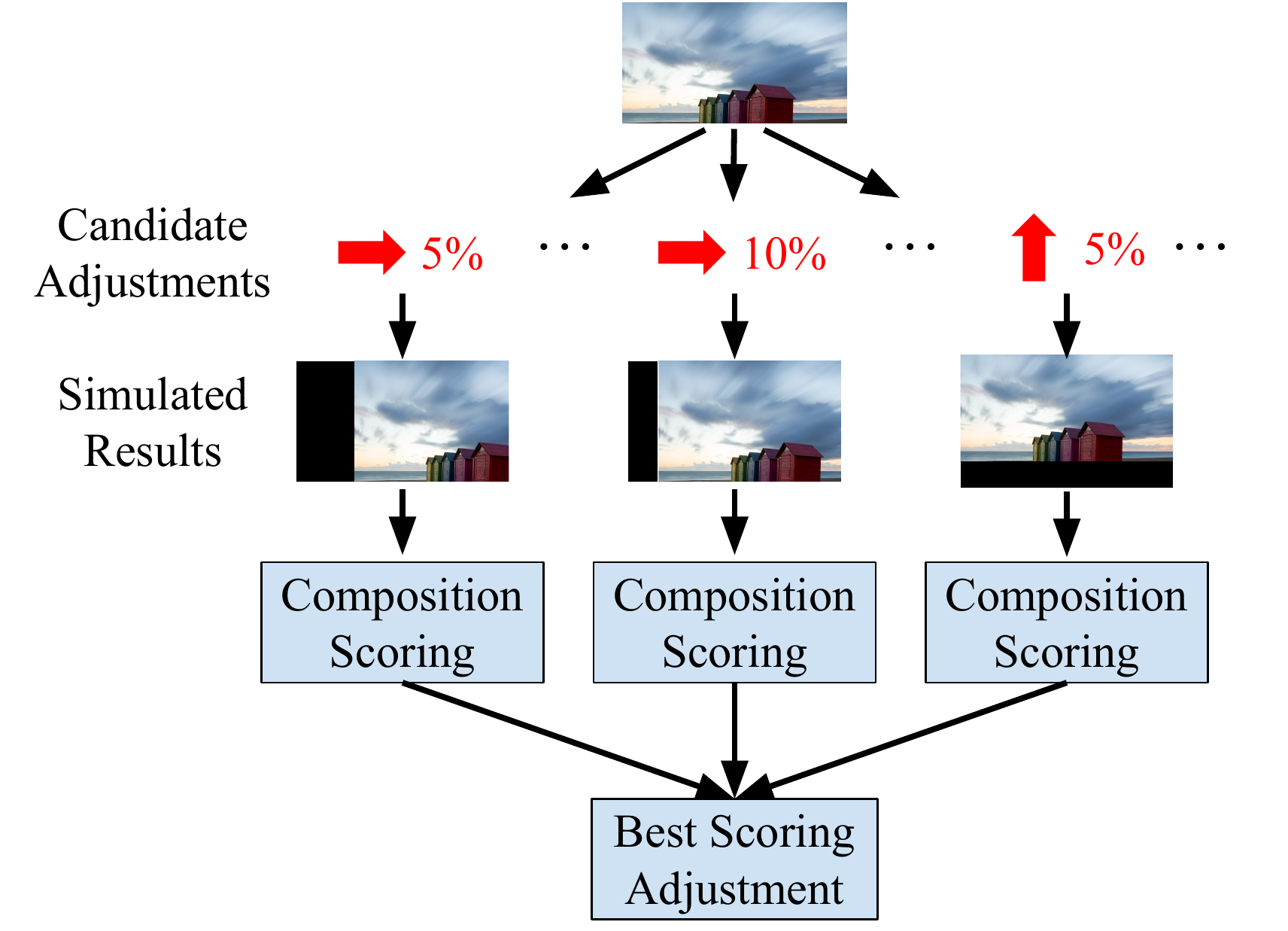}
    \vspace{-15pt}
    \caption{
        Pseudo label generation.
        If the composition score improvement is below a given threshold,
        the label is set to \emph{no adjustment is needed}.
        Otherwise, we use the best scoring adjustment as the label.
    }
    \label{fig:pseudo_label}
    \vspace{-15pt}
\end{figure}

\subsection{Pseudo Label Generation}
\vspace*{-0.01in}
\label{sec:pseudo_labels}

We generate pseudo view adjustment labels using the composition scoring model through simulation.
Given an image, we simulate the results of candidate view adjustments.
For each candidate adjustment, we generate the results for nine magnitudes equally spaced between $[\frac{\pi}{36}, \frac{\pi}{4}]$ for rotation and $[5, 45]$ for other adjustments (represented as the percentage of the image size).
This leads to $8{\times}9{=}72$ different view adjustment results.
We use the composition scoring model to rate these candidate views and select the one with the highest score.
If the best view leads to a composition score higher than the original image by a margin of $\Delta{=}0.2$,
the adjustment is taken as the label.
Otherwise, the label is that no adjustment is required.
Similar to the unlabeled image perturbation in the composition scoring model,
unknown regions that come into the image during simulation are filled with zero pixels.
Note that we ignore depths when simulating the adjustment results, because depth information is generally unavailable.
See Fig.~\ref{fig:pseudo_label}.

\subsection{View Adjustment Model}
\label{sec:view_prediction_model}
\vspace{-0.05pt}

Given the labeled samples from the view adjustment dataset and the images with pseudo label generated by the composition scoring model,
we learn a view adjustment model as follows.
The view adjustment model is implemented using a CNN with three output heads:
\begin{itemize}[leftmargin=*,label=$\bullet$,topsep=4pt]
    \setlength{\itemsep}{2pt}
    \setlength{\parskip}{2pt}
    \item \emph{Suggestion predictor} predicts whether a view adjustment should be suggested.
    \emph{Suggestion predictor} is a binary classification head trained with cross-entropy loss.
    \item \emph{Adjustment predictor} predicts which candidate view adjustment should be suggested when adjustment is needed.
    It is a multi-class classification head trained with categorical cross-entropy loss.
    \item \emph{Magnitude predictor} predicts the adjustment magnitude given the suggested adjustment.
    The \emph{magnitude predictor} consists of eight regressors that regress the adjustment magnitude for each of the candidate adjustments and is trained with $\ell_1$ loss.
\end{itemize}
During training, the gradients of the adjustment predictor are backpropagated only for samples where a suggestion should be provided.
Similarly, the gradients of the magnitude predictor are only backpropagated for the adjustment that should be suggested.

During inference, we first use the suggestion predictor to decide whether or not the view needs to be adjusted.
If the view needs to be adjusted, we use the adjustment predictor to select the candidate adjustment.
Finally, we use the output of the magnitude predictor corresponding to the selected adjustment as the suggested magnitude.

We also examined other model designs, e.g.~combining the suggestion and adjustment predictors as a single multi-class classifier,
combining the adjustment and magnitude predictors, etc.
However, these alternative models do not work as well as the proposed method.
In particular, they fail significantly in suggestion prediction, i.e.~determining whether an adjustment is needed.

\section{Experiments}
\vspace*{-0.05in}
\label{sec:experiments}
We evaluate our view adjustment model both objectively and subjectively.
The goal is to verify that 1) our two-stage training approach improves the view adjustment prediction performance,
and 2) the suggested view adjustments improve the image composition.

\vspace*{-0.1in}
\paragraph{Training datasets}
As described in Sec.~\ref{sec:approach},
the proposed approach exploits an array of datasets for training.\\[4pt]
\emph{FCDB}~\cite{chen-wacv2017} is an image cropping dataset consisting of 1,265 training images annotated with best crops.\\[4pt]
\emph{GAICD}~\cite{zhang2019deep} is an image cropping dataset consisting of 1,036 training images.
Each image comes with 90 candidate crops along with their composition scores, and we take the one with the highest score as the best crop.\\[4pt]
\noindent \emph{CPC}~\cite{wei2018good} consists of 10,797 images. Each image comes with 24 candidate crops and their composition scores.\\[4pt]
\emph{Unsplash}~\cite{unsplashdata} consists of 2M images shared by 200K photographers on \href{https://unsplash.com/data}{unsplash.com}.
Since many of these images are contributed by experienced photographers,
we assume that they are well-composed and use them as unlabeled samples for training the composition scoring model.\\[5pt]
\emph{Open Images}~\cite{openimages} consists of 9M images from \href{https://www.flickr.com/}{Flickr}.
We generate pseudo labels for a subset of 5.5M images\footnote{We use images containing human verified labels.}.\\[-5pt]

For the composition scoring model,
we use the annotated best crops in \emph{FCDB} and \emph{GAICD} (with $K{=}16$),
the scored candidate crops in \emph{CPC} and \emph{GAICD} (with $N{=}16$),
and the unlabeled images in \emph{Unsplash} (with $P{=}16$).
For the view adjustment model,
we use labeled samples from \emph{FCDB} and \emph{GAICD}, and pseudo-labeled samples from \emph{Open Images}. In each training iteration, we use 64 labeled and 64 pseudo-labeled samples.

\vspace*{-0.1in}
\paragraph{Comparative methods}
Since there is no prior published work on view adjustment models, we use the following variants of our approach for comparison.
The goal is to demonstrate the effectiveness of the proposed two-stage semi-supervised training approach.
\begin{itemize}[leftmargin=*,label=$\bullet$,topsep=2pt]
    \setlength{\itemsep}{2pt}
    \setlength{\parskip}{2pt}
    \item \textbf{Supervised} --  The view adjustment model is trained using only the labeled data from the view adjustment dataset. The unlabeled images from \emph{Open Images} are not used.
    \item \textbf{Aesthetic-scoring} -- Motivated by existing image cropping approaches~\cite{wang2017deep,deng2018aesthetic}, we use aesthetic scores to train the image scoring model. Specifically, we train the scoring model to regresses the mean opinion scores of 250K images from the \emph{AVA}~\cite{MurrayMP12} dataset, which is a widely-used image aesthetics dataset.
    \item \textbf{Supervised-scoring} -- The composition scoring model is trained using only labeled data from FCDB, GAICD and CPC. The unlabeled images from \emph{Unsplash} are not used.
\end{itemize}
The network architecture and training details are shared across all methods.

\vspace*{-0.1in}
\paragraph{Implementation details}
Both the view adjustment and composition scoring models use MobileNet~\cite{HowardZCKWWAA17} architecture with input size $299{\times}299$.
The last convolution layer is followed by a spatial pyramid pooling layer ($1{\times}1$, $2{\times}2$ and $5{\times}5$)~\cite{HeZR014} and two fully-connected layers with 1,024 units and ReLU activation.
The composition scoring model has an output layer with a single node that uses the sigmoid activation function,
and the view adjustment prediction model has an output layer with three heads as described in Sec.~\ref{sec:view_prediction_model}.

All models are trained asynchronously on 5 P100 GPUs starting from ImageNet~\cite{imagenet} pretrained weights.
We use the ADAM~\cite{adam} optimizer with a learning rate of $2e^{-5}$ and weight decay of $5e^{-4}$.
The composition scoring models are trained for 250K steps and the view adjustment models are trained for 480K steps.
Because the distribution of view adjustments in the wild is unknown, we give equal importance to all adjustments and weight each sample by the inverse frequency of the adjustment predictor label.

\subsection{Evaluation Dataset and Metrics}
\vspace*{-0.05in}
\label{sec:eval_datasets}

A major contribution of this work is introducing a benchmark dataset for evaluating view adjustment models.
This section describes the evaluation dataset and the proposed evaluation metrics.

The proposed view adjustment evaluation dataset is created from the test splits of image cropping datasets \emph{FCDB} and \emph{GAICD}, following the procedure described in Sec.~\ref{sub:problem_formulation}.
For each image, we generate an evaluation sample for each candidate view adjustment by randomly sampling a magnitude in the range $[5, 45]$ for shift and zoom adjustment and $[\frac{\pi}{36}, \frac{\pi}{4}]$ for rotation.
See Table~\ref{tab:eval_dataset} for statistics.
Note that the possible perturbations for an image are determined by the size of its best crop.
For example, horizontal shift and zoom-out are not possible for a crop whose width is same as the image width. Therefore, each candidate view adjustments has a different number of evaluation samples.

\begin{table}[t]\small
    \centering
    \tabcolsep=0.12cm
    \begin{tabular}{lllll}
        Horizontal & Vertical & Zoom & Rotate & Total \\
        \midrule
        258 (Left)  & 370 (Up)   & 268 (In)  & 255 (Clockwise) & 3,076\\
        277 (Right) & 350 (Down) & 521 (Out) & 256 (Counter)   \\
        \bottomrule
    \end{tabular}
    \vspace{-8pt}
    \caption{Number of evaluations samples for each candidate view adjustment.}
    \label{tab:eval_dataset}
    \vspace{-12pt}
\end{table}

\begin{table*}[t]
\vspace{-4pt}
\small
\centering
\begin{tabular}{lccccccccccc}
      &     &     & \multicolumn{8}{c}{F1 score} & \\
      \cmidrule{4-11}
      & AUC & TPR & Left & Right & Up & Down & Zoom-in & Zoom-out & Clockwise & Counter & IoU\\
      \midrule
    Supervised         & 0.570 & 0.366 & 0.126 & 0.074 & 0.0 & 0.0 & 0.022 & 0.364 & 0.082 & 0.061 & 0.730 \\
    Aesthetic-scoring  & 0.536 & 0.352 & 0.052 & 0.115 & 0.0 & 0.0 & \textbf{0.102} & 0.302 & 0.061 & 0.088 & 0.717 \\
    Supervised-scoring & 0.536 & 0.356 & 0.078 & 0.082 & 0.164 & 0.234 & 0.007 & 0.350 & 0.118 & \textbf{0.117} & 0.737 \\
    Proposed           & \textbf{0.608} & \textbf{0.436} & \textbf{0.221} & \textbf{0.221} & \textbf{0.390} & \textbf{0.341} & 0.015 & \textbf{0.378} & \textbf{0.124} & 0.110 & \textbf{0.750} \\
\bottomrule
\end{tabular}
\vspace{-8pt}
\caption{Performance of view adjustment models trained with different strategies. The TPR, IoU and F1 scores are measured at 0.3 FPR for the suggestion predictor.}
\label{tab:results}
\vspace{-12pt}
\end{table*}

We propose the following objective metrics for evaluating view adjustment models.
The goal is to evaluate 1) how accurately does a model trigger suggestions,
2) whether the suggested adjustment is correct when the model provides a suggestion,
and 3) how well the suggested view approximates the best view.
\begin{itemize}[leftmargin=*,label=$\bullet$,topsep=3pt]
    \setlength{\itemsep}{2pt}
    \setlength{\parskip}{1pt}
    \item \emph{Area under receiver operating characteristics curve (AUC)} measures the performance of the suggestion predictor, i.e., measures how accurately a model triggers suggestions.
    \item \emph{True positive rate (TPR)} also measures the performance of the suggestion predictor.
    More specifically, we measure the TPR at $0.3$ False Positive Rate (FPR).
    We choose a low FPR because suggesting an adjustment when it is not needed has a higher cost (asking the user to do unnecessary work and making the composition worse) compared to not suggesting an adjustment when the composition can be improved.
    All other metrics are computed at the same $0.3$ FPR operating point.
    \item \emph{F1-score} for each candidate adjustment measures the accuracy of the adjustment predictor and how accurate we are on the suggested adjustment.
    \item \emph{Intersection over Union (IoU)} between the predicted and ground truth views measures how well the suggested view approximates the best view.
    It evaluates the performance of the suggestion predictor, adjustment predictor, and magnitude predictor jointly.
\end{itemize}
Besides the objective metrics, we also conduct subjective evaluation through a user study.

\begin{figure}[t]
    \vspace{-3pt}
    \centering
    \includegraphics[width=.95\linewidth]{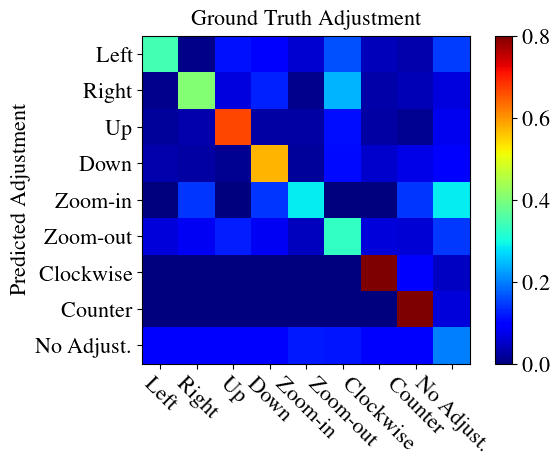}
    \vspace{-15pt}
    \caption{
        Confusion matrix.
        We normalize along each row, so each cell shows the precision of adjustment prediction.
    }
    \label{fig:confusion}
    \vspace{-12pt}
\end{figure}

\subsection{Objective Evaluation}

Table~\ref{tab:results} summarizes the performance of various methods on the view adjustment evaluation dataset.
Overall, the proposed semi-supervised approach achieves the best performance when compared with supervised alternatives.
Our model clearly outperforms supervised approaches in determining whether a suggestion should be provided.
It achieves $43.6\%$ TPR when the FPR is $30\%$.
Similarly, our model achieves the best F1-score for all candidate adjustments except for the \emph{Zoom-in} and \emph{Counter} clockwise rotation adjustment.
The absolute F1-score for \emph{Zoom-in} and \emph{Rotation} adjustments are low because of low recall values.
This may be explained by the adjustment magnitudes distribution,
where most zoom-in and rotation samples have small perturbation magnitudes and are by nature more difficult for view adjustment prediction.
Finally, our method achieves the best IoU, which means that the suggested views best approximate the ground truth views.
We also compare with off-the-shelf cropping models~\cite{wei2018good,zeng2020cropping} on our view adjustment dataset,
but their performance is significantly worse than our method. The IoU between predicted crops and ground truth views is below 0.61.
The results justify that cropping is not sufficient for solving the view adjustment problem.

To further understand when our model fails,
we show the confusion matrix in Fig.~\ref{fig:confusion}.
The results show that our model rarely makes the mistake of choosing the wrong shift or rotation adjustment.
Instead, most of the errors occur between zoom-out and shifting.
This is not surprising,
because both zoom-out and shifting can resolve the problem where the original view wrongly excludes important content.

Comparing the performance of our model with the supervised alternatives,
we can clearly see that the two-stage approach helps to improve model performance.
The results also show that semi-supervised learning is important not only for the view adjustment model but also for the composition scoring model.
If we train the composition scoring model using only supervised data,
the results are actually worse than the purely supervised view adjustment model.
This is potentially due to the poor quality of pseudo labels generated by the supervised composition scoring model,
which is trained on a limited number of images.
Finally, the performance of \textbf{Aesthetic-scoring} is worse than other methods.
This suggests that a generic aesthetic score is not sufficient to differentiate the changes caused by camera view adjustment.

\begin{table}[t]\small
    \tabcolsep=0.12cm
    \vspace{-12pt}
    \centering
    \begin{tabular}{lccc}
                                   & After is better & Before is better & No difference \\
        \midrule
        View Adjustment            & $79.0\%$ & $12.6\%$ & $8.5\%$ \\
        Pano2Vid~\cite{su2016accv} & $78.9\%$ & $17.1\%$ & $4.1\%$ \\
    \bottomrule
    \end{tabular}
    \vspace{-8pt}
    \caption{
        Subjective evaluation results on the view adjustment and the Pano2Vid dataset.
        The raters are asked whether each image has better composition before or after applying the suggested view adjustment.
    }
    \vspace{-12pt}
    \label{tab:subjective}
\end{table}

\begin{figure}[t]
    \vspace{-8pt}
    \centering
    \includegraphics[width=\linewidth]{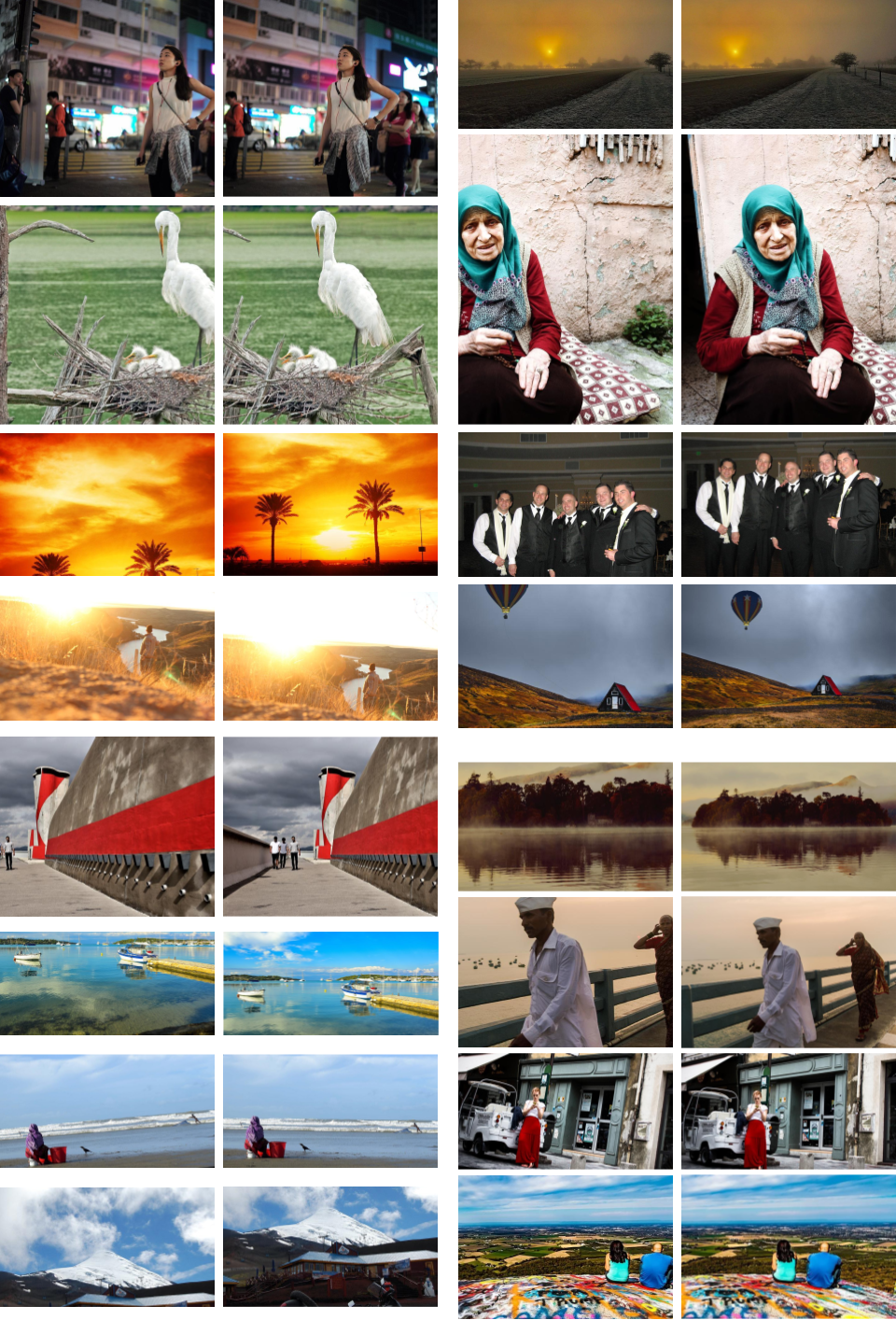}
    \vspace{-18pt}
    \caption{
        Qualitative examples.
        Each pair shows the original image (left) and the result of the adjustment (right).
        Our model appears to provide suggestions based on various factors, including objects, the ground plane, or even leading lines in the image.
        The last two rows show examples where our model predicts the wrong adjustment but still improves the image composition.
    }
    \label{fig:qualitative_results}
    \vspace{-9pt}
\end{figure}

\subsection{Subjective Evaluation}
\label{sub:subjective}

\begin{figure}[t]
    \vspace{-8pt}
    \centering
    \begin{subfigure}[t]{0.5\linewidth}
        \centering
        \includegraphics[width=\linewidth]{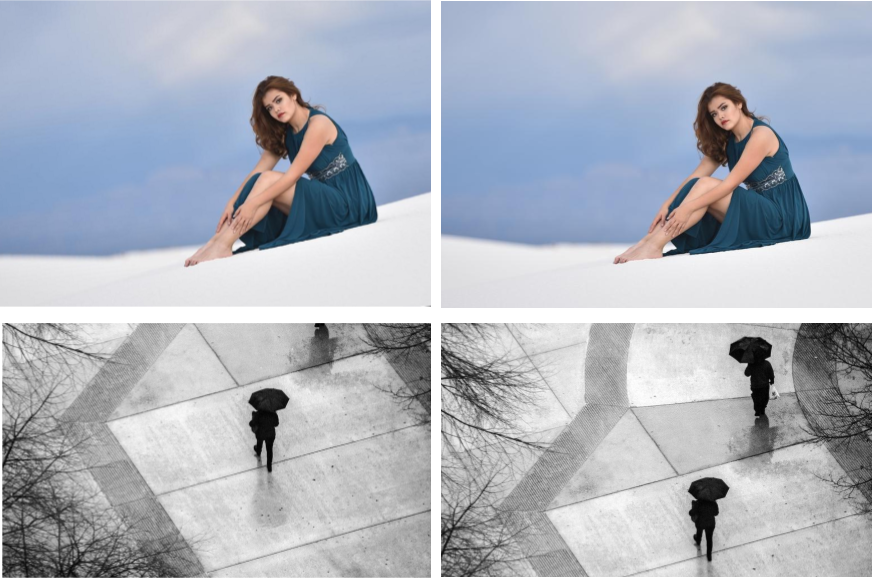}
        \vspace{-15pt}
        \caption{
            False negatives.
        }
        \label{fig:withinimage_objectsize}
    \end{subfigure}
    \begin{subfigure}[t]{0.46\linewidth}
        \centering
        \includegraphics[width=\linewidth]{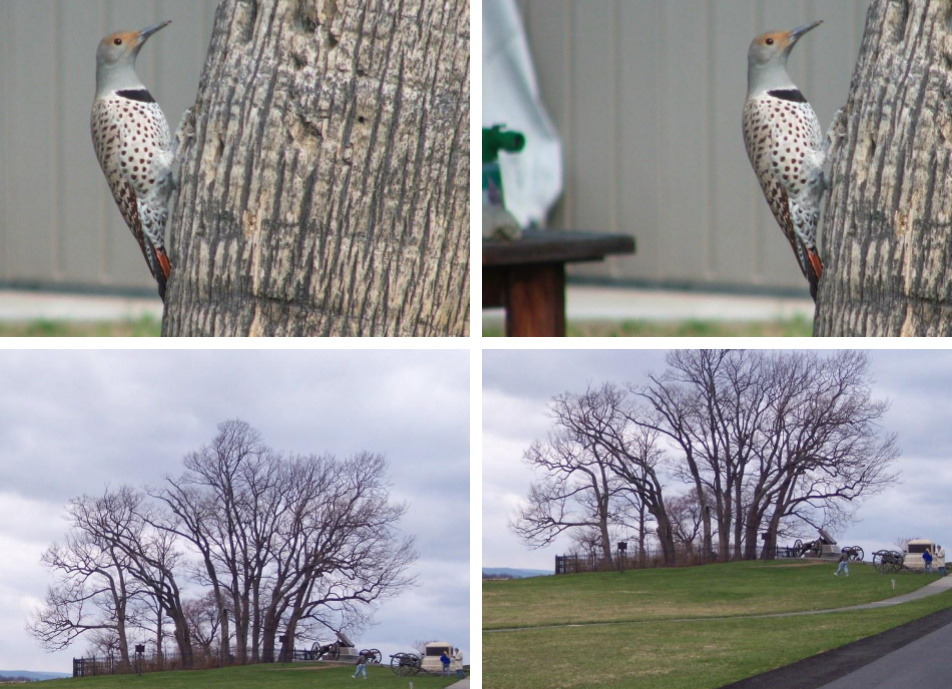}
        \vspace{-15pt}
        \caption{
            False positives.
        }
        \label{fig:withinimage_objectsize}
    \end{subfigure}
    \vspace{-9pt}
    \caption{
        Failure examples.
        False negative means that our model fails to provide a suggestion when the composition can be improved. Each pair shows the input image (left) and the ground truth (right).
        False positive means the suggested adjustment degrades the composition according to our user study. Each pair shows the input image (left) and our prediction (right).
        Our model fails when the target adjustment is minor or the important content is not visible in the initial view.
        Also, the bottom right example shows how a wrong magnitude prediction degrades the composition. 
    }
    \label{fig:failure_cases}
    \vspace{-3pt}
\end{figure}

To demonstrate the effectiveness of the proposed model,
we conduct a user study that directly evaluates whether the suggested view adjustment improves the composition of the original image.
In the user study,
we show the raters the image both before and after applying the suggested view adjustment.
The raters are asked which image has the better composition or if they cannot tell.
The order of the two images is chosen randomly to avoid bias.
We select $130$ images from our evaluation dataset for the user study,
and each image is rated by 3 different raters.
The results are in Table~\ref{tab:subjective}.
It shows that our model has a high success rate ($79.0\%$) when a suggestion is provided,
and it only wrongly suggests an adjustment about $12.6\%$ of the time.
See Fig.~\ref{fig:qualitative_results} and Fig.~\ref{fig:failure_cases} for qualitative examples.

Beside the view adjustment dataset,
we also conduct a user study on the \emph{Pano2Vid}~\cite{su2016accv} dataset.
\emph{Pano2Vid} is a $360\degree$ video dataset with viewport annotations.
The viewports capture a normal FOV video created from the $360\degree$ video by raters and cover interesting content with appropriate composition.
We sample $200$ frames from the \emph{Pano2Vid} dataset and apply our model on the user-selected viewport.
The horizontal and vertical adjustments become translations in spherical coordinates, and zoom-in and zoom-out adjustments become changes in the FOV.
We run the model iteratively until the model stops suggesting any adjustment or reaches the maximum number of steps (3 steps),
and we ask raters whether the final view has a better composition than the initial view.

\begin{figure}[t]
    \centering
    \includegraphics[width=\linewidth]{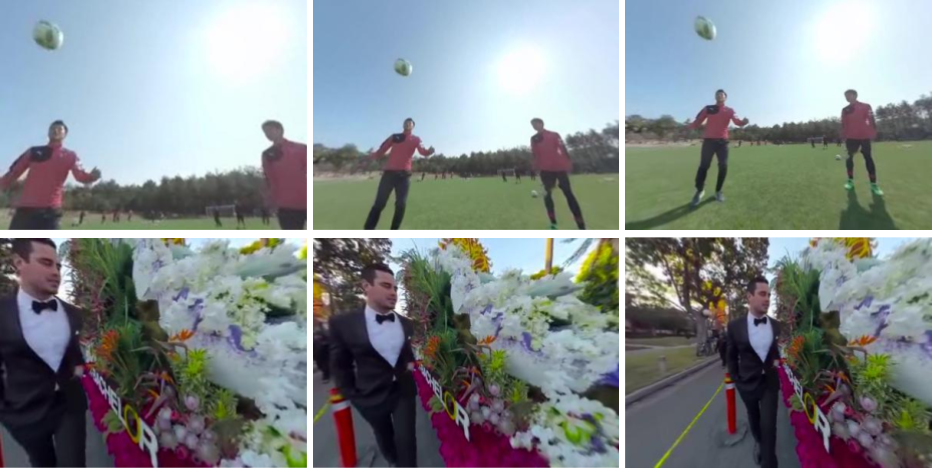}
    \vspace{-15pt}
    \caption{
        Qualitative examples from $360\degree$ images.
        We perform multi-step view adjustment on $360\degree$ images,
        and our model improves the composition iteratively.
    }
    \label{fig:qualitative_pano2vid}
    \vspace{-12pt}
\end{figure}

The results on \emph{Pano2Vid} are in Table~\ref{tab:subjective}.
Interestingly, the success rate of the model remains high but has a slightly higher failure rate.
The results show that our model generalizes well despite the fact that it has never been trained on $360\degree$ images.
See Fig.~\ref{fig:qualitative_pano2vid} for examples.
Note that we perform multi-step adjustment on \emph{Pano2Vid},
so the results also imply that our model can improve the composition iteratively even if it is trained only for single step view adjustment.
This is important in practice,
because the best view may not be reachable from the initial view using a single adjustment when applying the system in the wild.
On the other hand,
we also observe that our model may drift far from the initial view in some examples.
In other words, it suggests a final view that is unrelated to the initial view,
which shows a limitation of the single-step adjustment model.

\vspace*{-0.05in}
\section{Conclusion}
\vspace*{-0.05in}
We propose improving image composition through view adjustment prediction.
Our system suggests how to improve the composition while the user is still composing the photo.
We create a new benchmark dataset for evaluating the performance of a view adjustment prediction and propose a two-stage training approach that exploits both labeled and unlabeled data for learning the view adjustment model.
Experimental results show that the suggestion provided by our model improves image composition $79\%$ of the time.
Future plans are to explore the iterative nature of view adjustment for more accurate and consistent suggestions.

{\small
%\bibliographystyle{ieee_fullname}
%\bibliography{references}

\begin{thebibliography}{10}\itemsep=-1pt

\bibitem{camera51}
Camera51--a smarter camera.
\newblock
  \url{https://play.google.com/store/apps/details?id=com.camera51.android}.
\newblock Accessed: 2020-09-22.

\bibitem{unsplashdata}
Unsplash dataset.
\newblock \url{https://unsplash.com/data}.

\bibitem{samsung-shot-suggestion}
What is shot suggestions on galaxy?
\newblock
  \url{https://www.samsung.com/global/galaxy/what-is/shot-suggestions/}.
\newblock Accessed: 2020-09-22.

\bibitem{chen2016automatic}
Jiansheng Chen, Gaocheng Bai, Shaoheng Liang, and Zhengqin Li.
\newblock Automatic image cropping: A computational complexity study.
\newblock In {\em CVPR}, 2016.

\bibitem{chen2003visual}
Li-Qun Chen, Xing Xie, Xin Fan, Wei-Ying Ma, Hong-Jiang Zhang, and He-Qin Zhou.
\newblock A visual attention model for adapting images on small displays.
\newblock {\em Multimedia systems}, 9(4):353--364, 2003.

\bibitem{chen-wacv2017}
Yi-Ling Chen, Tzu-Wei Huang, Kai-Han Chang, Yu-Chen Tsai, Hwann-Tzong Chen, and
  Bing-Yu Chen.
\newblock Quantitative analysis of automatic image cropping algorithms:a
  dataset and comparative study.
\newblock In {\em WACV}, 2017.

\bibitem{chen-acmmm-2017}
Yi-Ling Chen, Jan Klopp, Min Sun, Shao-Yi Chien, and Kwan-Liu Ma.
\newblock Learning to compose with professional photographs on the web.
\newblock In {\em ACM Multimedia}, 2017.

\bibitem{cheng2010learning}
Bin Cheng, Bingbing Ni, Shuicheng Yan, and Qi Tian.
\newblock Learning to photograph.
\newblock In {\em ACM Multimedia}, 2010.

\bibitem{imagenet}
Jia Deng, Wei Dong, Richard Socher, Li{-}Jia Li, Kai Li, and Fei{-}Fei Li.
\newblock Imagenet: {A} large-scale hierarchical image database.
\newblock In {\em CVPR}, 2009.

\bibitem{deng2018aesthetic}
Yubin Deng, Chen~Change Loy, and Xiaoou Tang.
\newblock Aesthetic-driven image enhancement by adversarial learning.
\newblock In {\em ACM Multimedia}, 2018.

\bibitem{fang2014automatic}
Chen Fang, Zhe Lin, Radomir Mech, and Xiaohui Shen.
\newblock Automatic image cropping using visual composition, boundary
  simplicity and content preservation models.
\newblock In {\em ACM Multimedia}, 2014.

\bibitem{fang2017creatism}
Hui Fang and Meng Zhang.
\newblock Creatism: A deep-learning photographer capable of creating
  professional work.
\newblock {\em arXiv preprint arXiv:1707.03491}, 2017.

\bibitem{HeZR014}
Kaiming He, Xiangyu Zhang, Shaoqing Ren, and Jian Sun.
\newblock Spatial pyramid pooling in deep convolutional networks for visual
  recognition.
\newblock In David~J. Fleet, Tom{\'{a}}s Pajdla, Bernt Schiele, and Tinne
  Tuytelaars, editors, {\em ECCV}, 2014.

\bibitem{HowardZCKWWAA17}
Andrew~G. Howard, Menglong Zhu, Bo Chen, Dmitry Kalenichenko, Weijun Wang,
  Tobias Weyand, Marco Andreetto, and Hartwig Adam.
\newblock Mobilenets: Efficient convolutional neural networks for mobile vision
  applications.
\newblock {\em CoRR}, abs/1704.04861, 2017.

\bibitem{hu2019virtualportraitist}
Chuan-Shen Hu, Yi-Tsung Hsieh, Hsiao-Wei Lin, and Mei-Chen Yeh.
\newblock Virtual portraitist: An intelligent tool for taking well-posed
  selfies.
\newblock {\em ACM Trans. Multimedia Comput. Commun. Appl.}, 15(1), Jan. 2019.

\bibitem{huang2012mm}
Yen-Ta Huang, Kuan-Ting Chen, Liang-Chi Hsieh, Winston Hsu, and Ya-Fan Su.
\newblock Detecting the directions of viewing landmarks for recommendation by
  large-scale user-contributed photos.
\newblock In {\em ACM Multimedia}, 2012.

\bibitem{adam}
Diederik~P. Kingma and Jimmy Ba.
\newblock Adam: {A} method for stochastic optimization.
\newblock In {\em ICLR}, 2015.

\bibitem{openimages}
Ivan Krasin, Tom Duerig, Neil Alldrin, Vittorio Ferrari, Sami Abu-El-Haija,
  Alina Kuznetsova, Hassan Rom, Jasper Uijlings, Stefan Popov, Andreas Veit,
  Serge Belongie, Victor Gomes, Abhinav Gupta, Chen Sun, Gal Chechik, David
  Cai, Zheyun Feng, Dhyanesh Narayanan, and Kevin Murphy.
\newblock Openimages: A public dataset for large-scale multi-label and
  multi-class image classification.
\newblock {\em Dataset available from https://github.com/openimages}, 2017.

\bibitem{li2018a2}
Debang Li, Huikai Wu, Junge Zhang, and Kaiqi Huang.
\newblock A2-rl: Aesthetics aware reinforcement learning for image cropping.
\newblock In {\em CVPR}, 2018.

\bibitem{li2019fast}
Debang Li, Huikai Wu, Junge Zhang, and Kaiqi Huang.
\newblock Fast a3rl: Aesthetics-aware adversarial reinforcement learning for
  image cropping.
\newblock {\em IEEE Transactions on Image Processing}, 28(10):5105--5120, 2019.

\bibitem{Li_2020_CVPR}
Debang Li, Junge Zhang, and Kaiqi Huang.
\newblock Learning to learn cropping models for different aspect ratio
  requirements.
\newblock In {\em CVPR}, 2020.

\bibitem{Li_2020_CVPR_Mutual}
Debang Li, Junge Zhang, Kaiqi Huang, and Ming-Hsuan Yang.
\newblock Composing good shots by exploiting mutual relations.
\newblock In {\em CVPR}, 2020.

\bibitem{liu2010optimizing}
Ligang Liu, Renjie Chen, Lior Wolf, and Daniel Cohen-Or.
\newblock Optimizing photo composition.
\newblock {\em Computer Graphics Forum}, 29(2):469--478, 2010.

\bibitem{lu2020weakly}
Peng Lu, Jiahui Liu, Xujun Peng, and Xiaojie Wang.
\newblock Weakly supervised real-time image cropping based on aesthetic
  distributions.
\newblock In {\em ACM Multimedia}, 2020.

\bibitem{lu2019end}
Peng Lu, Hao Zhang, Xujun Peng, and Xiaofu Jin.
\newblock An end-to-end neural network for image cropping by learning
  composition from aesthetic photos.
\newblock {\em arXiv preprint arXiv:1907.01432}, 2019.

\bibitem{lu2019listwise}
Weirui Lu, Xiaofen Xing, Bolun Cai, and Xiangmin Xu.
\newblock Listwise view ranking for image cropping.
\newblock {\em IEEE Access}, 7:91904--91911, 2019.

\bibitem{ma2014icip}
Shuang Ma, Yangyu Fan, and Chang~Wen Chen.
\newblock Finding your spot: A photography suggestion system for placing human
  in the scene.
\newblock In {\em ICIP}, 2014.

\bibitem{ma2014pose}
Shuang Ma, Yangyu Fan, and Chang~Wen Chen.
\newblock Pose maker: A pose recommendation system for person in the landscape
  photographing.
\newblock In {\em ACM Multimedia}, 2014.

\bibitem{marchesotti2009framework}
Luca Marchesotti, Claudio Cifarelli, and Gabriela Csurka.
\newblock A framework for visual saliency detection with applications to image
  thumbnailing.
\newblock In {\em ICCV}, 2009.

\bibitem{MurrayMP12}
Naila Murray, Luca Marchesotti, and Florent Perronnin.
\newblock {AVA:} {A} large-scale database for aesthetic visual analysis.
\newblock In {\em CVPR}, 2012.

\bibitem{nishiyama2009sensation}
Masashi Nishiyama, Takahiro Okabe, Yoichi Sato, and Imari Sato.
\newblock Sensation-based photo cropping.
\newblock In {\em ACM Multimedia}, 2009.

\bibitem{rawat2015context}
Yogesh~Singh Rawat and Mohan~S Kankanhalli.
\newblock Context-aware photography learning for smart mobile devices.
\newblock {\em ACM Transactions on Multimedia Computing, Communications, and
  Applications}, 12(1s):1--24, 2015.

\bibitem{rawat2016clicksmart}
Yogesh~Singh Rawat and Mohan~S Kankanhalli.
\newblock Clicksmart: A context-aware viewpoint recommendation system for
  mobile photography.
\newblock {\em IEEE Transactions on Circuits and Systems for Video Technology},
  27(1):149--158, 2016.

\bibitem{rawat2018tmm}
Y.~S. {Rawat}, M. {Song}, and M.~S. {Kankanhalli}.
\newblock A spring-electric graph model for socialized group photography.
\newblock {\em IEEE Transactions on Multimedia}, 20(3):754--766, 2018.

\bibitem{stentiford2007attention}
Fred Stentiford.
\newblock Attention based auto image cropping.
\newblock In {\em International Conference on Computer Vision Systems:
  Proceedings (2007)}, 2007.

\bibitem{su2012tmm}
Hsiao-Hang Su, Tse-Wei Chen, Chieh-Chi Kao, Winston~H. Hsu, and Shao-Yi Chien.
\newblock Preference-aware view recommendation system for scenic photos based
  on bag-of-aesthetics-preserving features.
\newblock {\em IEEE Transactions on Multimedia}, 14(3):833--843, 2012.

\bibitem{su2016accv}
Yu-Chuan Su, Dinesh Jayaraman, and Kristen Grauman.
\newblock Pano2vid: Automatic cinematography for watching 360$\degree$ videos.
\newblock In {\em ACCV}, 2016.

\bibitem{suh2003automatic}
Bongwon Suh, Haibin Ling, Benjamin~B Bederson, and David~W Jacobs.
\newblock Automatic thumbnail cropping and its effectiveness.
\newblock In {\em UIST}, 2003.

\bibitem{abs-1911-10492}
Yi Tu, Li Niu, Weijie Zhao, Dawei Cheng, and Liqing Zhang.
\newblock Image cropping with composition and saliency aware aesthetic score
  map.
\newblock In {\em AAAI}, 2020.

\bibitem{wang2017deep}
Wenguan Wang and Jianbing Shen.
\newblock Deep cropping via attention box prediction and aesthetics assessment.
\newblock In {\em ICCV}, 2017.

\bibitem{wang2015tist}
Yinting Wang, Mingli Song, Dacheng Tao, Yong Rui, Jiajun Bu, Ah~Chung Tsoi,
  Shaojie Zhuo, and Ping Tan.
\newblock Where2stand: A human position recommendation system for souvenir
  photography.
\newblock {\em ACM Trans. Intell. Syst. Technol.}, 7(1), Oct. 2015.

\bibitem{wei2018good}
Zijun Wei, Jianming Zhang, Xiaohui Shen, Zhe Lin, Radomir Mech, Minh Hoai, and
  Dimitris Samaras.
\newblock Good view hunting: Learning photo composition from dense view pairs.
\newblock In {\em CVPR}, 2018.

\bibitem{xu2014should}
Pengfei Xu, Hongxun Yao, Rongrong Ji, Xian-Ming Liu, and Xiaoshuai Sun.
\newblock Where should i stand? learning based human position recommendation
  for mobile photographing.
\newblock {\em Multimedia tools and applications}, 69(1):3--29, 2014.

\bibitem{yan2013learning}
Jianzhou Yan, Stephen Lin, Sing Bing~Kang, and Xiaoou Tang.
\newblock Learning the change for automatic image cropping.
\newblock In {\em CVPR}, 2013.

\bibitem{yin2012crowdsourced}
Wenyuan Yin, Tao Mei, and Chang~Wen Chen.
\newblock Crowdsourced learning to photograph via mobile devices.
\newblock In {\em ICME}, 2012.

\bibitem{yin2013socialized}
Wenyuan Yin, Tao Mei, Chang~Wen Chen, and Shipeng Li.
\newblock Socialized mobile photography: Learning to photograph with social
  context via mobile devices.
\newblock {\em IEEE Transactions on Multimedia}, 16(1):184--200, 2013.

\bibitem{zhang2019deep}
Hui Zeng, Lida Li, Zisheng Cao, and Lei Zhang.
\newblock Reliable and efficient image cropping: A grid anchor based approach.
\newblock In {\em CVPR}, 2019.

\bibitem{zeng2020cropping}
Hui Zeng, Lida Li, Zisheng Cao, and Lei Zhang.
\newblock Grid anchor based image cropping: A new benchmark and an efficient
  model.
\newblock {\em IEEE Transactions on Pattern Analysis and Machine Intelligence},
  2020.

\bibitem{zhang2005auto}
Mingju Zhang, Lei Zhang, Yanfeng Sun, Lin Feng, and Weiying Ma.
\newblock Auto cropping for digital photographs.
\newblock In {\em ICME}, 2005.

\bibitem{zhang2012icip}
Yanhao Zhang, Xiaoshuai Sun, Hongxun Yao, Lei Qin, and Qingming Huang.
\newblock Aesthetic composition representation for portrait photographing
  recommendation.
\newblock In {\em ICIP}, 2012.

\bibitem{zhang2013location}
Y. {Zhang} and R. {Zimmermann}.
\newblock Camera shooting location recommendations for landmarks in geo-space.
\newblock In {\em International Symposium on Modelling, Analysis and Simulation
  of Computer and Telecommunication Systems}, 2013.

\end{thebibliography}

}

\pagebreak
\section*{Appendices}
\label{sec:appendix}
\renewcommand{\thesubsection}{\Alph{subsection}}

The supplementary materials consist of:
\begin{enumerate}[leftmargin=*,label=\Alph*]
    \item Perturbation details for the composition scoring model
    \item Perturbation details for the view adjustment model
    \item Data augmentation details
    \item Training pipeline overview
    \item Experiment results analysis
    \item Additional qualitative examples
\end{enumerate}

\subsection{Perturbation for Composition Scoring}

In this section, we describe the perturbation details for generating the image pairs $(I_{p}, I_{n})$ for training the composition scoring model.

Let $c_{x}$, $c_{y}$, $w$ and $h$ denote the center $x$, $y$ coordinate, width and height of the best crop bounding box normalized to $[0, 1]$ for each image.
Given the perturbation ($o_{x}$, $o_{y}$, $o_{z}$, $o_{\alpha}$) where $o_{x}$, $o_{y}$, $o_{z}$, and $o_{\alpha}$ are the magnitudes of horizontal shift, vertical shift, zoom, and rotation respectively,
the perturbed bounding box is defined as
\begin{equation}
    \label{eq:bbox_perturbation}
    \begin{aligned}
        c^{\prime}_{x} &= c_{x} + w * o_{x}\\
        c^{\prime}_{y} &= c_{y} + h * o_{y}\\
        w^{\prime} &= w + w * o_{z}\\
        h^{\prime} &= h + h * o_{z}\\
        \alpha &= o_{\alpha}.
    \end{aligned}
\end{equation}
Note that $\alpha$ is the orientation of the bounding box in the image coordinate system and is defined as the counter clockwise angle between the $y$-axis of the image and the bounding box.
Therefore, the coordinates of the bounding box corners in the image are given by $u = c + R_{\alpha} v$, where $R_{\alpha}$ is the rotation matrix and $v = [\pm \frac{w}{2}, \pm \frac{h}{2}]$.

Specifically, we apply four types of perturbation:
\begin{itemize}[leftmargin=*,label=$\bullet$]
    \setlength{\itemsep}{1pt}
    \setlength{\parskip}{1pt}
    \item \textbf{Shifting} randomly samples $o_{x}$ and $o_{y}$ in the range $[-0.4, 0.4]$ and applies $(o_{x}, o_{y}, 0, 0)$ to the bounding box.
    \item \textbf{Zooming-out} randomly samples $o_{z}$ in the range $[0, 0.4]$ and applies $(0, 0, o_{z}, 0)$ to the bounding box.
    \item \textbf{Cropping} combines zooming-in and shifting.
    It first samples $o_{z}$ in the range $[\sqrt{0.5}, \sqrt{0.8}]$ and then samples $o_{x}$ and $o_{y}$ in the range $[-o_{z}/2, o_{z}/2]$.
    The perturbation $(o_{x}, o_{y}, o_{z}, 0)$ is then applied to the bounding box.
    In other words, it randomly sample a crop within the original crop where the area is 0.5 to 0.8 times that of the original one and the aspect ratio is the same.
    \item \textbf{Rotation} randomly samples $o_{\alpha}$ in the range $[-\frac{\pi}{4}, \frac{\pi}{4}]$ and applies $(0, 0, 0, o_{
    \alpha})$ to the bounding box.
\end{itemize}
We apply the same perturbations for both the labeled crops and unlabeled images.
For unlabeled image, the bounding box is set to the entire image,
i.e.~$c_{x}=c_{y}=0.5$, and $w=h=1.0$.
Also, as described in the main paper,
we discard samples where the perturbed bounding box goes beyond the original image for supervised samples.

\subsection{Perturbation for View Adjustment}

This section describes how we generate samples for view adjustment prediction.
The generated samples are used for both training and evaluation.

Given an image crop represented as $(c_{x}, c_{y}, w, h)$,
we apply the following perturbations:
\begin{itemize}[leftmargin=*,label=$\bullet$]
    \setlength{\itemsep}{1pt}
    \setlength{\parskip}{1pt}
    \item \textbf{Horizontal shift} samples $o_{x}$ in either $[0.05, 0.45]$ or $[-0.45, -0.05]$ and applies perturbation $(o_{x}, 0, 0, 0)$ to the bounding box.
    \item \textbf{Vertical shift} samples $o_{y}$ in either $[0.05, 0.45]$ or $[-0.45, -0.05]$ and applies perturbation $(0, o_{y}, 0, 0)$ to the bounding box.
    \item \textbf{Zoom} samples $o_{z}$ in either $[-0.048, -0.310]$ or $[0.053, 0.818]$ and applies $(0, 0, o_{z}, 0)$ to the bounding box.
    \item \textbf{Rotation} samples $o_{\alpha}$ in either $[-\frac{\pi}{4}, -\frac{\pi}{36}]$ or $[\frac{\pi}{36}, \frac{\pi}{4}]$ and applies $(0, 0, 0, o_{\alpha})$ to the bounding box.
\end{itemize}
Note that the ranges for shift and zoom perturbations are determined such that the inverse perturbation,
i.e., the ground truth view adjustment, falls in the range of $[0.05, 0.45]$.
As described in the main paper, we discard samples where the perturbed bounding box falls outside the image.

\begin{algorithm}[t]
    \SetAlgoLined
    \DontPrintSemicolon
    \KwResult{Composition scoring model $M_{c}$}
    \For{$i \gets 1, \text{MaxStep}$} {
        \tcp{Scored candidate samples}
        $I_{s}\,, crop_{s} \gets$ $N$ images with scored crops\;
        $crop_{p}, crop_{n} \gets$ create $\frac{N\times(N-1)}{2}$ pairs of crops\;
        $I^{p}_{s} \gets$ ExtractCrop($I_{s}$, $crop_{p}$)\;
        $I^{n}_{s} \gets$ ExtractCrop($I_{s}$, $crop_{n}$)\;
        \tcp{Best crop samples}
        $I_{c}\,, crop_{c} \gets$ $K$ images with best crop\;
        $crop^{\prime}_{c} \gets$ Perturb($crop_{c}$)\;
        $I^{p}_{c} \gets$ ExtractCrop($I_{c}$, $crop_{c}$)\;
        $I^{n}_{c} \gets$ ExtractCrop($I_{c}$, $crop^{\prime}_{c}$)\;
        \tcp{Unsupervised samples}
        $I^{p}_{u} \gets$ $P$ well-composed images\;
        $I^{n}_{u} \gets$ Perturb($I^{p}_{u}$)\;
        \tcp{Data augmentation}
        $I^{i}_{j} \gets$ Augment($I^{i}_{j}$) $\forall (i, j) \in \{p, n\} \times \{s, c\}$\;
        $I^{p}_{u} \gets$ Augment($I^{p}_{u}$)\;
        \tcp{Minimize joint loss}
        $y^{p} \gets M_{c}(I^{p})$\;
        $y^{n} \gets M_{c}(I^{n})$\;
        $M_{c} \gets \argmin_{M_{c}} \, \sum_{x} Loss(y^{p}_{x}, y^{n}_{x})$
    }
    \caption{Composition scoring model training}
    \label{alg:composition}
\end{algorithm}

\subsection{Data Augmentation}

This section describes the implementation details for the composition scoring model data augmentation.
For \emph{shift borders} augmentation, we randomly select a percentage $s_{y}$ in $[0, 40]$ and select $s_{x}\%$ of columns on either the left or right side of the image and replace the pixel values with zero.
Similarly, we randomly select $s_{y}\%$ of rows from either the top or bottom of the image and replace the pixel values.
Note that $s_{x}$ and $s_{y}$ are samples independently.
For \emph{zoom-out} augmentation, we randomly select a percentage $s_{z}$ in $[0, 40]$ and select $0.5s_{z}\%$ of rows and columns on both sides of the image.
For \emph{rotation} augmentation, we randomly select an angle $\theta \in [-\frac{\pi}{4}, \frac{\pi}{4}]$ and apply two rotation operator $R_{-\theta} \circ R_{\theta}$ to the image sequentially.
Note that while $R_{-\theta} \circ R_{\theta} = I$ mathematically, if we apply the two operator sequentially on an image with finite size, they introduce pixels not within the original image and therefore simulate the synthetic artifacts introduced by rotation.

\subsection{Training Pipeline Overview}

This section provides an overview of the training pipeline.
See Algorithm~\ref{alg:composition} and Algorithm~\ref{alg:adjustment} for the training of the composition scoring and view adjustment model respectively.

\begin{algorithm}[t]
    \SetAlgoLined
    \DontPrintSemicolon
    \KwResult{View adjustment model $M_{v}$}
    \For{$i \gets 1, \text{MaxStep}$} {
        \tcp{Pseudo-labeled samples}
        $I_{s} \gets$ $K$ images\;
        $y_{s} \gets$ PseudoLabel($I_{s}$)\;
        \tcp{Labeled samples}
        $I_{c}\,, crop_{c} \gets$ $K$ images with best crop\;
        $p_{c} \gets$ samples random perturbations\; 
        $crop^{\prime}_{c} \gets$ Perturb($crop_{c}$, $p_{c}$)\;
        $I^{\prime}_{c} \gets$ ExtractCrop($I_{c}$, $crop^{\prime}_{c}$)\;
        $y_{c} \gets p^{-1}_{c}$\;
        \tcp{Minimize joint loss}
        $y^{\prime}_{s} \gets M_{v}(I_{s})$\;
        $y^{\prime}_{c} \gets M_{v}(I^{\prime}_{c})$\;
        $M_{v} \gets \argmin_{M_{v}} \, Loss(y_{s}, y^{\prime}_{s}) + Loss(y_{c}, y^{\prime}_{c})$
    }
    \caption{View adjustment model training}
    \label{alg:adjustment}
\end{algorithm}

\subsection{Experiment Results}

\begin{table}[t]\small
    \tabcolsep=0.12cm
    \centering
    \begin{tabular}{lccc}
                   & After is better & Before is better & No difference \\
        \midrule
        Horizontal & $80.0\%$ & $5.0\%$ & $15.0\%$ \\
        Vertical   & $94.2\%$ & $0.0\%$ & $5.8\%$ \\
        Zoom       & $74.7\%$ & $16.0\%$ & $9.3\%$ \\
        Rotation   & $57.3\%$ & $41.3\%$ & $1.3\%$ \\
        \midrule
        Single-step & $81.6\%$ & $14.9\%$ & $3.5\%$\\
        Multi-steps & $72.2\%$ & $22.2\%$ & $5.6\%$\\
    \bottomrule
    \end{tabular}
    \vspace{-6pt}
    \caption{
        Subjective evaluation results on the view adjustment (row1--row4) and the Pano2Vid (row5--row6) dataset.
    }
    \label{tab:subjective}
\end{table}

In this section, we expand the experiment results in the main paper by providing more detailed analysis on the subjective evaluation results.
In particular, we analyze the user study results w.r.t.~1) different view adjustments being suggested on the view adjustment dataset,
and 2) the number of adjustment being performed on the Pano2Vid dataset.

The results are in Table~\ref{tab:subjective}.
Interestingly, the subjective evaluation results have a strong dependency on the adjustment being suggested.
Our method achieves the best performance in vertical shift adjustment, where it improves the composition more than $94\%$ of the time without any false positive.
In contrast, the rotation suggestion only improves the composition $57.3\%$ of the time.
This is primarily caused by the incorrect magnitude prediction, because image composition is very sensitive to the orientation when the horizon is visible.
Therefore, even a small error in magnitude prediction may significantly degrade the result.

Not surprisingly, our method has a better performance in single-step view adjustment.
In contrast, the success rate drops by $9.4\%$ when multiple adjustments are required.
The results show the need for considering the iterative nature of view adjustment in future works.

\subsection{Qualitative Results}

\begin{figure*}[t]
    \includegraphics[width=\linewidth]{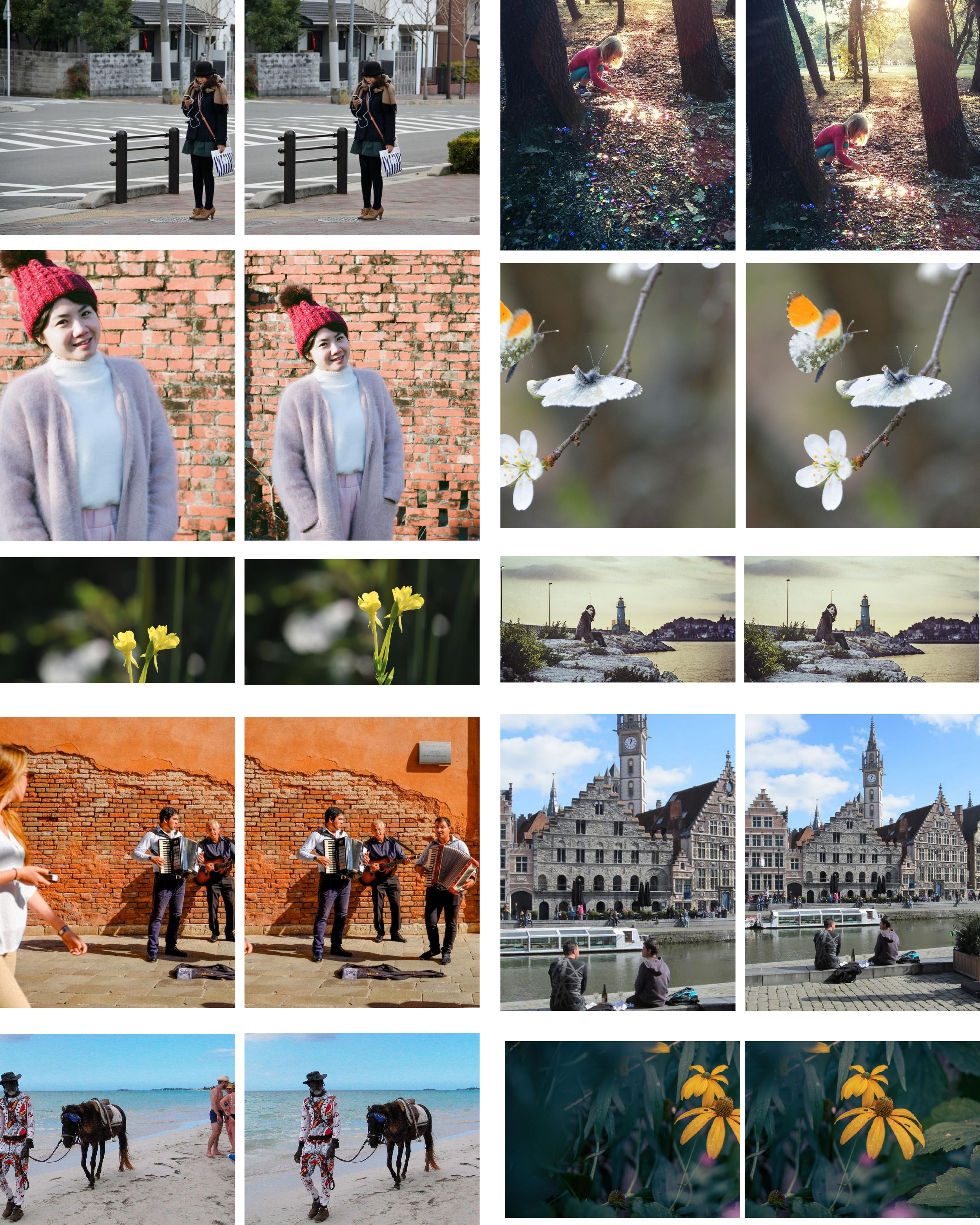}
    \caption{
        Qualitative examples.
        Each pair shows the original image (left) and our prediction (right).
    }
    \label{fig:supp_qualitative}
\end{figure*}

\begin{figure*}[t]
    \includegraphics[width=\linewidth]{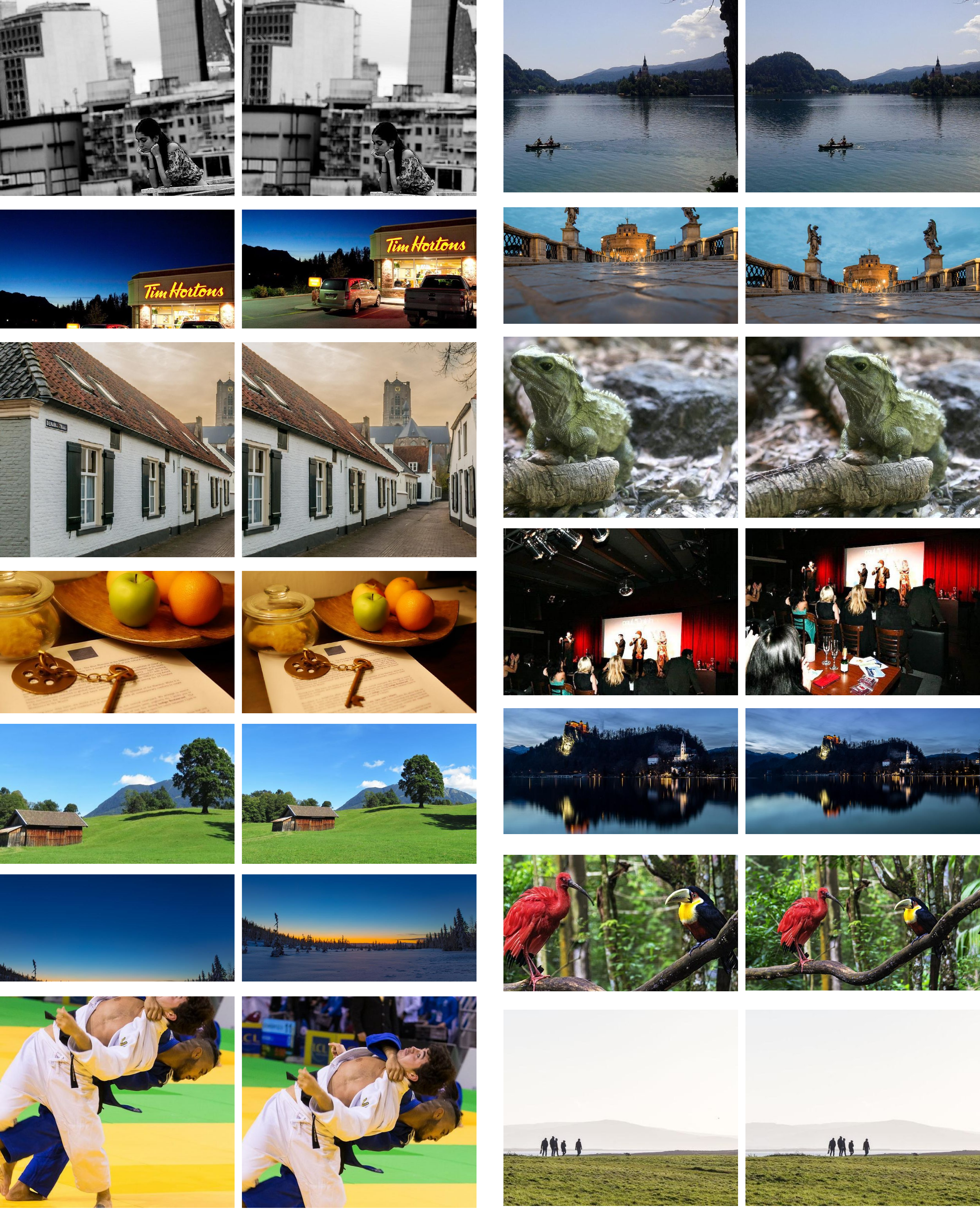}
    \caption{
        Qualitative examples (cont.).
        Each pair shows the ground truth (left) and our prediction (right).
    }
    \label{fig:supp_qualitative_2}
\end{figure*}

In this section, we show additional qualitative examples similar to Fig.~6 in the main paper.
See Fig.~\ref{fig:supp_qualitative} and Fig.~\ref{fig:supp_qualitative_2}.
The results show that our model works in a wide range of scenarios, including both object centric and scenic photos.
Also, our model works for different aspect ratios, including photos in both landscape and portrait orientations.

\end{document}